\documentclass[conference]{IEEEtran}
\IEEEoverridecommandlockouts

\usepackage{cite}
\usepackage{amsmath,amssymb,amsfonts}
\usepackage{algorithmic}
\usepackage{graphicx}
\usepackage{booktabs}
\usepackage{textcomp}
\usepackage{xcolor}
\def\BibTeX{{\rm B\kern-.05em{\sc i\kern-.025em b}\kern-.08em
    T\kern-.1667em\lower.7ex\hbox{E}\kern-.125emX}}
\begin{document}

\title{Geometry-Free Conditional Diffusion Modeling for Solving the Inverse Electrocardiography Problem
}

\author{\IEEEauthorblockN{Ramiro Valdes Jara}
\IEEEauthorblockA{\textit{Department of Industrial and Systems Engineering} \\
\textit{University of Miami}\\
Coral Gables, United States of America \\
rjv71@miami.edu}
\and
\IEEEauthorblockN{Adam Meyers}
\IEEEauthorblockA{\textit{Department of Industrial and Systems Engineering} \\
\textit{University of Miami}\\
Coral Gables, United States of America \\
axm8336@miami.edu}
}

\maketitle

\begin{abstract}
This paper proposes a data-driven model for solving the inverse problem of electrocardiography, the mathematical problem that forms the basis of electrocardiographic imaging (ECGI). We present a conditional diffusion framework that learns a probabilistic mapping from noisy body surface signals to heart surface electric potentials. The proposed approach leverages the generative nature of diffusion models to capture the non-unique and underdetermined nature of the ECGI inverse problem, enabling probabilistic sampling of multiple reconstructions rather than a single deterministic estimate. Unlike traditional methods, the proposed framework is geometry-free and purely data-driven, alleviating the need for patient-specific mesh construction. We evaluate the method on a real ECGI dataset and compare it against strong deterministic baselines, including a convolutional neural network, long short-term memory network, and transformer-based model. The results demonstrate that the proposed diffusion approach achieves improved reconstruction accuracy, highlighting the potential of diffusion models as a robust tool for noninvasive cardiac electrophysiology imaging.
\end{abstract}

\begin{IEEEkeywords}
Inverse electrocardiography, diffusion model, data-driven, generative model, noninvasive cardiac imaging, body surface potentials, epicardial potentials
\end{IEEEkeywords}

\section{Introduction}
Cardiovascular diseases are among the leading causes of death \cite{LINDSTROM20222372}. Understanding the mechanisms underlying hearts requires high-resolution mapping of the electrical activity on the heart surface \cite{ref1highresolution}, \cite{ref2highresolution}. However, current clinical techniques for acquiring such information are invasive, limited to a single procedure and typically performed under sedation, rendering them impractical. As a result, clinicians often rely on noninvasive body surface recordings, which are easier to acquire but lack the high resolution needed to capture complex electrical patterns within the heart \cite{ref1invasive}, \cite{ref2invasive}. This gap shows the need for noninvasive and patient-friendly methods capable of providing detailed cardiac electrical information to support diagnosis and therapy planning.

Electrocardiography (ECG) measures electric potentials on the torso surface that arise from the heart's electrical activity. In computational terms, this relationship is often described through a \emph{forward problem} and an \emph{inverse problem}. The ECG forward problem aims to predict signals on the body surface from known cardiac activity, whereas the ECG inverse problem seeks to infer cardiac information from ECG or body surface potentials (BSPs). Electrocardiographic imaging (ECGI) refers to the noninvasive imaging framework that reconstructs cardiac electrical activity from body surface measurements, and in practice, it is formulated by solving the inverse problem of electrocardiography.

\begin{figure}[t]
    \centering
    \includegraphics[width=0.9\linewidth]{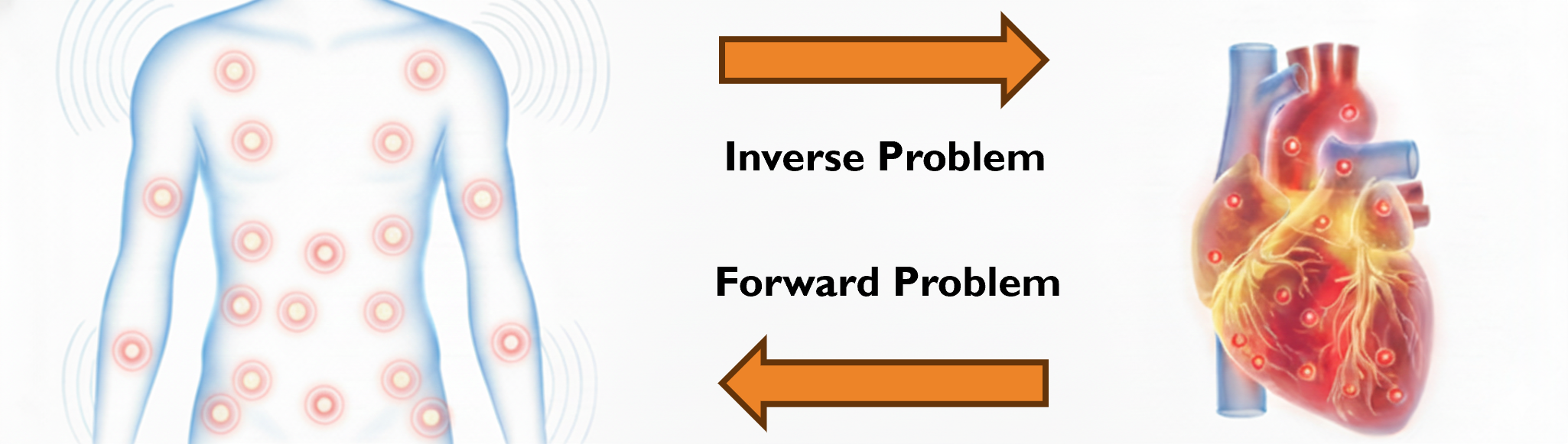}
    \caption{Illustration of the forward and inverse problems in electrocardiography.}
    \label{fig:ecg_forward_inverse}
\end{figure}

In a typical ECGI setting, the forward model can be expressed as a linear operator mapping cardiac sources to torso measurements (Fig.~\ref{fig:ecg_forward_inverse}) as
\begin{equation}
\mathbf{y}(t) = \mathbf{A}\,\mathbf{x}(t) + \boldsymbol{\eta}(t),
\label{eq:forward}
\end{equation}
where $\mathbf{x}(t)$ represents the cardiac electrical quantity of interest (e.g. heart surface potentials), $\mathbf{y}(t)$ is the BSP/ECG measurement vector, $\mathbf{A}$ is the transfer matrix (often derived from solving the Laplace equation in a patient-specific geometry), and $\boldsymbol{\eta}(t)$ denotes measurement noise and modeling error.

The ECGI inverse problem attempts to recover $\mathbf{x}(t)$ from $\mathbf{y}(t)$ given a patient's geometry. However, this inversion is inherently ill-posed and underdetermined, as multiple distinct cardiac electrical configurations can produce similar body surface potentials. This lack of uniqueness is further compounded by measurement noise and modeling inaccuracies, such that small perturbations in $\mathbf{y}(t)$ can lead to large variations in the reconstructed $\mathbf{x}(t)$. Consequently, ECGI solutions incorporate priors, assumptions about what constitutes a plausible heart surface electrical signal. In classical methods, these priors are typically enforced through regularization terms that promote smoothness, whereas more recent data-driven approaches learn such priors directly from training data.

This study proposes a data-driven conditional diffusion model for solving the ECGI problem within a probabilistic generative framework. The model learns a conditional distribution that maps noisy body surface potential measurements to heart surface electric potentials. Unlike deterministic approaches that yield a single point estimate, the diffusion model accounts for the non-uniqueness inherent in the inverse problem by modeling a distribution of feasible cardiac reconstructions consistent with the observed BSP signals. The method is geometry-free and purely data-driven, alleviating the need for explicit anatomical modeling. We evaluate the proposed approach on the task of reconstructing heart surface potentials from noisy BSP measurements and compare it against strong deterministic baselines to assess improvements in accuracy and robustness.


\section{Literature Review}

Methods for solving the inverse problem of electrocardiography can be 
classified into deterministic approaches and probabilistic approaches \cite{li2024solvinginverseproblemelectrocardiography}.
Deterministic methods typically formulate ECGI as a regularized optimization
problem and produce a single solution, while probabilistic methods aim to
model uncertainty and incorporate prior knowledge through learned or
statistical distributions. In recent years, data-driven
and deep learning techniques have been introduced within both paradigms.

\subsection{Deterministic approaches}
Classical ECGI methods formulate the inverse problem as the minimization of a
data fidelity term combined with regularization. A common formulation
is given by
\begin{equation}
\hat{\mathbf{x}} = \arg\min_{\mathbf{x}} 
\; \|\mathbf{y} - \mathbf{A}\mathbf{x}\|_2^2 
\;+\; \lambda^2 \|\mathbf{\Gamma}\mathbf{x}\|_2^2 ,
\label{eq:tikhonov}
\end{equation}
where $\Gamma$ denotes a regularization operator and $\lambda$ a regularization parameter. Among these methods, Tikhonov regularization is the most widely adopted in
ECGI. Zero-order Tikhonov regularization ($\Gamma=\mathbf{I}$) penalizes large
solution amplitudes, while first order variants enforce spatial smoothness by
penalizing gradients on the heart surface  \cite{Ramanathan2004ECGI}, \cite{9120232}, \cite{5075589}, \cite{schuler}. Several related deterministic techniques have also been explored. Truncated singular value decomposition (TSVD) mitigates noise amplification by discarding
small singular values of the forward matrix \cite{5478914}. Truncated total least squares
(TTLS) extends this idea by accounting for perturbations in both measurements
and the forward operator \cite{4384223}. Other approaches incorporate $\ell_1$-norm or total
variation (TV) regularization to promote sparsity \cite{6817555}. Furthermore, to account for the temporal nature of cardiac activity,
spatiotemporal regularization methods extend these formulations by coupling
solutions across time, such as Twomey regularization \cite{1673624}. Classical deterministic approaches are attractive due to their
explicit use of the forward model. However, typical priors imposed by standard regularization are often insufficient to capture the complex spatiotemporal structure of cardiac electrical activity, particularly
in highly noisy settings.

Recent advances in deep learning have motivated data-driven approaches. Such methods aim to learn a direct mapping from body surface
measurements to cardiac electrical quantities using paired datasets, in which torso potentials and their corresponding epicardial signals are available. Early studies
applied fully connected or convolutional neural networks (CNNs) to localize
arrhythmic sources or reconstruct cardiac activation patterns from ECG signals \cite{cnnecgi}, \cite{cnnecgiv2}. More recent work has incorporated geometric information
through graph-based neural networks, where the heart surface is represented as
a mesh or graph and convolution is performed over its topology
\cite{8049505}, \cite{meister2020graphconvolutionalregressioncardiac}, \cite{10.1007/978-3-030-87231-1_53}, \cite{9932432}. Temporal dependencies have been incorporated using recurrent architectures, such as
long short-term memory, or by explicitly feeding multiple time steps into the network
\cite{Chen2022ML_ECGI}. Compared to classical regularization methods, neural
networks can learn richer spatial and temporal patterns directly from the data.

Despite their potential, purely data-driven approaches may require large and diverse training datasets to generalize well, and without explicit constraints, they may
produce solutions that are not fully consistent with the underlying physiology governing the heart. To address these limitations, physics-informed approaches have been
proposed. These methods incorporate physical constraints, such as the forward
operator or governing equations \cite{Ugurlu_pulse}. Nevertheless, physics-informed models also have practical drawbacks. They require accurate knowledge of the underlying physical model and its parameters (e.g., patient geometry or tissue properties), which are often unavailable in practice. In addition, model training can be sensitive to the choice of loss weighting and hyperparameters.

Overall, although deterministic methods provide data-consistent solutions, they rely on fixed regularization assumptions and produce only a single point estimate. As a result, they may struggle in highly noisy settings and do not explicitly quantify uncertainty, motivating the development of probabilistic approaches.

\subsection{Probabilistic approaches}

Probabilistic ECGI methods formulate the inverse problem within a Bayesian
framework, where cardiac electrical activity is treated as a random variable
and inference is performed by combining a likelihood derived from the forward
model with a prior distribution. Under Gaussian noise and Gaussian prior
assumptions, this leads to maximum a posteriori (MAP) estimators that are
closely related to classical Tikhonov regularization, providing a statistical
interpretation of deterministic regularization methods. Bayesian formulations based on Gaussian
processes have been proposed to impose spatiotemporal smoothness
while enabling uncertainty quantification. Sequential Bayesian approaches,
including Kalman filtering, have also been explored to model the
temporal evolution of cardiac electrical activity and explicitly account for
measurement noise \cite{1431075}, \cite{1703754}, \cite{10.1007/978-3-031-16434-7_42}.

Overall, probabilistic estimation approaches provide a way to
incorporate prior knowledge and quantify uncertainty in ECGI. However, existing
methods often rely on relatively simple priors (e.g., Gaussian or smoothness assumptions), which limits their ability
to model the high-dimensional, nonlinear spatiotemporal structure of cardiac electrical activity. This motivates the development of more expressive data-driven probabilistic models.

\subsection{Geometry-free inverse modeling}

Classical ECGI formulations rely on patient-specific anatomical models to construct a transfer matrix $A$, which encodes the physical relationship between cardiac electrical sources and body surface measurements. While this physics-based formulation provides strong data consistency, obtaining patient geometry requires dedicated imaging such as CT or MRI. In practice, these requirements significantly limit clinical scalability and restrict the applicability of ECGI.

Motivated by these limitations, an emerging line of research investigates geometry-free approaches, where the inverse mapping from surface measurements to cardiac electrical activity is learned directly from data, without explicitly constructing the transfer matrix $A$. This approach can be interpreted as learning a data-driven surrogate of the inverse operator, which shifts the problem from physics-constrained inversion to statistical inference. These approaches do not explicitly enforce the forward physical relationship and do not require patient-specific geometry. While such models may not satisfy strict data consistency constraints, they provide an alternative to classical ECGI, particularly in scenarios where anatomical information is unavailable or otherwise challenging to collect.

Several neural network-based models have been proposed to reconstruct cardiac electrical signals directly from body surface measurements. Ornelas-Mu\~{n}oz et al.~\cite{10981210} employ transformer architectures to map ECG signals to transmembrane voltage potentials, demonstrating that deep sequence models can recover high-dimensional cardiac signals without explicit geometric information. However, such transformer-based models are trained as deterministic point predictors, which can be limiting in highly ill-posed or noisy ECGI settings. Similarly, Guo et al.~\cite{diff_ecgi} introduce a diffusion-based framework to learn the inverse relationship between cardiac electrophysiology signals and ECG measurements, formulating the problem probabilistically and enabling uncertainty-aware reconstructions. Nevertheless, the approach of~\cite{diff_ecgi} has primarily been demonstrated on synthetic one-dimensional cardiac cable data and relies on guided noise initialization rather than explicit conditional diffusion on body surface measurements throughout the reverse process, potentially limiting the method's ability to capture high-dimensional spatiotemporal cardiac dynamics and long-range temporal dependencies.

This work adopts the perspective of focusing on probabilistic reconstruction of cardiac electrical activity from surface measurements without assuming access to patient geometry. By combining a transformer-based denoiser with conditional diffusion, we aim to obtain geometry-free reconstructions that can effectively capture high-dimensional spatiotemporal cardiac dynamics, addressing a key limitation of prior geometry-free methods.

\section{Methodology}

\begin{figure*}[t]
    \centering
    \includegraphics[width=\linewidth]{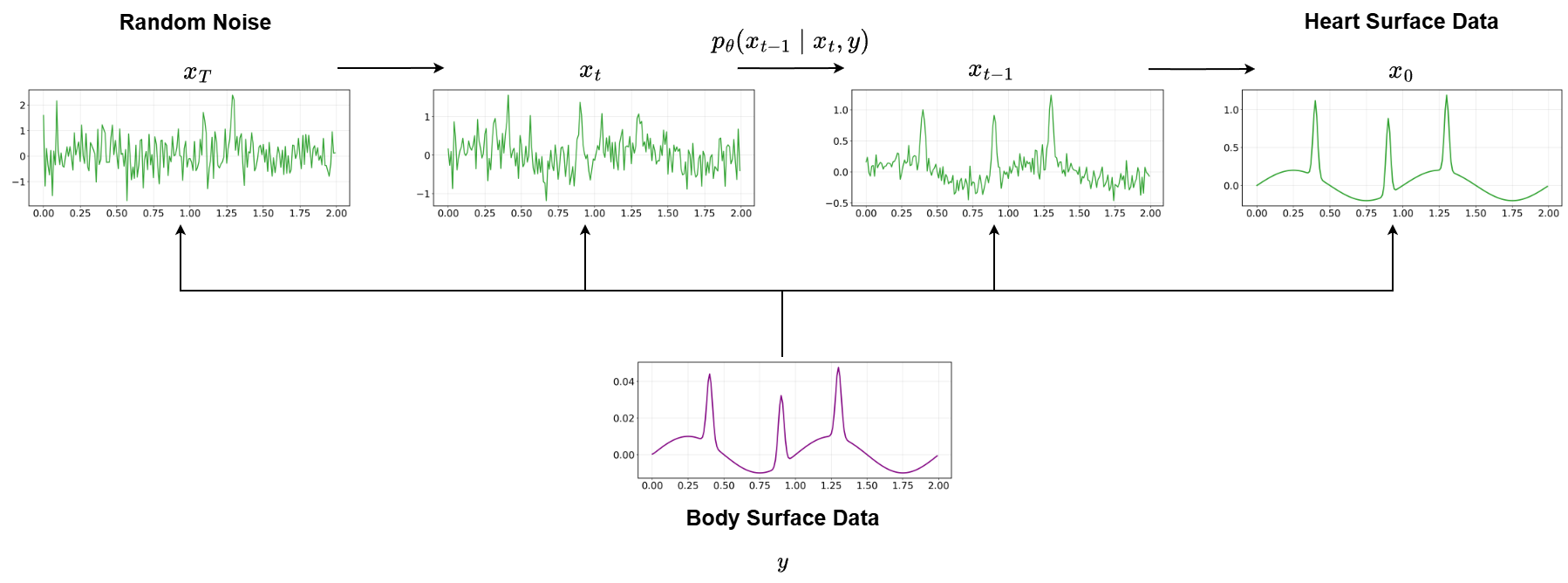}
    \caption{Scheme of our conditional diffusion model for ECGI reconstruction. 
The reverse diffusion process progressively denoises an initial random signal \(x_T\) to recover the cardiac surface potentials \(x_0\), guided by observed body surface measurements \(y\). 
At each timestep \(t\), the conditional transition \(p_\theta(x_{t-1} \mid x_t, y)\) incorporates the torso ECG to constrain the reconstruction toward physiologically consistent solutions.}
    \label{fig:diff_ecgi}
\end{figure*}

\subsection{Denoising diffusion probalistic model}
Let $\mathbf{x}_0 \in \mathbb{R}^{N \times T}$ denote a clean data sample. Consider the problem of learning a distribution $p_\theta(\mathbf{x}_0)$ that approximates the empirical data distribution $q(\mathbf{x}_0)$. 
To this end, we define a sequence of latent variables $\{\mathbf{x}_t\}_{t=1}^T$ that share the same domain as $\mathbf{x}_0$.
Diffusion models are latent variable models composed of two stochastic processes, including a forward (noising) process and a reverse (denoising) process. The forward process gradually corrupts the clean signal $x_0$ into the sequence of latent variables $\{\mathbf{x}_t\}_{t=1}^T$ by adding Gaussian noise according to a predefined variance schedule $\{\mathbf{\beta}_t\}_{t=1}^T$.

The forward process is defined as a Markov chain given by
\begin{equation}
q(x_t \mid x_{t-1}) = \mathcal{N}\!\left(x_t; \sqrt{1-\beta_t}\, x_{t-1}, \beta_t I \right),
\label{eq:forward_markov}
\end{equation}
where $\mathbf{\beta}_t \in (0,1)$ controls the noise level at timestep $t$. By composition of the transitions, $\mathbf{x}_t$ can be sampled directly from $\mathbf{x}_0$ in closed form as
\begin{equation}
x_t = \sqrt{\bar{\alpha}_t}\, x_0 + \sqrt{1-\bar{\alpha}_t}\,\epsilon, \qquad \epsilon \sim \mathcal{N}(0,I),
\label{eq:forward_closed}
\end{equation}
where $\alpha_t = 1-\beta_t$ and $\bar{\alpha}_t = \prod_{s=1}^t \alpha_s$. As $t$ increases, $\mathbf{x}_t$ approaches an isotropic Gaussian distribution. The reverse process seeks to progressively denoise $\mathbf{x}_t$ back to $\mathbf{x}_0$ and is parameterized as
\begin{equation}
p_\theta(x_{t-1} \mid x_t) = \mathcal{N}\!\big(x_{t-1}; \mu_\theta(x_t, t), \sigma_t^2 I \big),
\label{eq:reverse_general}
\end{equation}
where $\mu_\theta$ is a learnable mean function and $\sigma_t^2$ is typically fixed according to the forward schedule (we set $\sigma_t^2 = \beta_t$). Following the standard denoising diffusion probabilistic model (DDPM) parameterization introduced by Ho et al.~\cite{ho_diffusion}, the mean can be written as
\begin{equation}
\mu_\theta(x_t, t) = 
\frac{1}{\sqrt{\alpha_t}}\!\left(x_t - 
\frac{\beta_t}{\sqrt{1-\bar{\alpha}_t}}\, \epsilon_\theta(x_t, t) \right),
\label{eq:ddpm_mean}
\end{equation}
where $\epsilon_\theta$ is a neural network trained to predict the noise $\epsilon$ added in the forward process. Under this formulation, training reduces to minimizing a simple mean squared error loss between the true noise and the predicted noise, specifically
\begin{equation}
\mathcal{L}_{\text{diff}}(\theta) 
= \mathbb{E}_{x_0, \epsilon, t} 
\left[ \left\| \epsilon - \epsilon_\theta(x_t, t) \right\|_2^2 \right],
\label{eq:ddpm_loss}
\end{equation}
where $\mathbf{x}_t$ is generated according to Eq.~(\ref{eq:forward_closed}) and $t$ is sampled uniformly from $\{1,\dots,T\}$. During inference, sampling starts from $\mathbf{x}_T \sim \mathcal{N}(0,I)$ and iteratively applies Eq.~(\ref{eq:reverse_general}) until $\mathbf{x}_0$ is obtained, yielding samples from the learned data distribution.

\subsection{Conditional diffusion model for ECGI}
To solve the ECGI inverse problem, the above framework is extended to the conditional setting. We address the inverse problem of electrocardiography by modeling the
conditional distribution of cardiac electrical activity given body surface
measurements using a conditional diffusion model, illustrated in Fig.~\ref{fig:diff_ecgi}. Let $\mathbf{x}_0 \in
\mathbb{R}^{N_h \times T}$ denote the cardiac electrical activity 
(e.g., epicardial potentials over time) and let $\mathbf{y} \in
\mathbb{R}^{N_b \times T}$ denote the corresponding body surface potential
measurements. The relationship between $\mathbf{x}_0$ and $\mathbf{y}$ is governed by the ECG forward model defined in Eq. \eqref{eq:forward}. Our goal is to learn a probabilistic mapping from $\mathbf{y}$ to $\mathbf{x}_0$
by approximating the conditional distribution $p(\mathbf{x}_0 \mid \mathbf{y})$.
Rather than directly regressing a point estimate, we adopt a diffusion-based
generative approach that enables iterative refinement and probabilistic sampling of multiple reconstructions. In the conditional diffusion model, the reverse process is explicitly conditioned on the body surface measurements $\mathbf{y}$ as
\begin{equation}
p_\theta(x_{t-1} \mid x_t, y) 
= \mathcal{N}\!\big(x_{t-1}; \mu_\theta(x_t, t, y), \beta_t I \big).
\label{eq:cond_reverse}
\end{equation}

We again parameterize the mean via noise prediction
\begin{equation}
\mu_\theta(x_t, t, y) = 
\frac{1}{\sqrt{\alpha_t}}\!\left(x_t - 
\frac{\beta_t}{\sqrt{1-\bar{\alpha}_t}}\, \epsilon_\theta(x_t, t, y) \right),
\label{eq:cond_mean}
\end{equation}
where $\epsilon_\theta(x_t, t, y)$ denotes a learnable conditional noise-prediction function.

The corresponding training objective becomes
\begin{equation}
\mathcal{L}_{\text{cond}}(\theta) 
= \mathbb{E}_{x_0, y, \epsilon, t} 
\left[ \left\| \epsilon - \epsilon_\theta(x_t, t, y) \right\|_2^2 \right],
\label{eq:cond_loss}
\end{equation}

The denoising function $\epsilon_{\theta}(x_t, t, y)$ is implemented using a transformer-based architecture, with a design inspired by prior work on conditional diffusion models for time-series forecasting~\cite{rishi2025conditional}. Cardiac electrical activity exhibits strong temporal structure and nonlocal correlations across time, which can be challenging for convolutional or recurrent architectures to capture. Transformers are well suited for this task due to their ability to model long-range temporal dependencies and complex interactions in high-dimensional signals through self-attention, which allows the denoiser to adaptively focus on the most informative temporal regions when estimating the injected noise. At each diffusion timestep $t$, the network receives the noisy epicardial signal $\mathbf{x}_t$, the diffusion timestep $t$, and the body surface measurements $\mathbf{y}$ as conditional context. The network outputs an estimate of the added noise $\epsilon_{\theta}(x_t, t, y)$, which is then used to compute the reverse diffusion mean in Eq.~(\ref{eq:cond_mean}). 

This design allows the model to learn a rich data-driven prior over realistic cardiac electrical activity while implicitly capturing the structure of the ECG forward relationship present in the training data. During testing, samples from $p_\theta(x_0 \mid y)$ are generated by initializing $x_T \sim \mathcal{N}(0,I)$ and iteratively applying the conditional reverse process in Eq.~(\ref{eq:cond_reverse}) down to $t=0$. This procedure yields multiple plausible reconstructions of $x_0$ consistent with the observed BSP measurements $y$, providing a probabilistic solution to the ECGI inverse problem.

The objective of this paper is to demonstrate how conditional diffusion models can be
used to solve ill-posed inverse problems in cardiac electrophysiology. The forward diffusion process remains unchanged and consists of progressively adding Gaussian noise to the cardiac
signal, while the reverse process learns to denoise this signal under the guidance of the
observed BSP/ECG. Importantly, this approach does not incorporate a physics-based forward operator or patient-specific geometry; instead, the inverse relationship is learned directly from paired data.

\section{Experiments}

\subsection{Dataset Description}
The experiments in this work are conducted using the dataset
introduced by Ugurlu et al.~\cite{Ugurlu_pulse}, which consists of canine cardiac
electrophysiology recordings acquired in a controlled torso-tank setup at the University of Utah. In these experiments, isolated canine hearts are perfused and suspended in an electrolytic torso
tank under deep anesthesia. Epicardial electrical activity is recorded using a sock electrode
array containing 490 electrodes placed directly on the heart surface, with signals sampled at
1~kHz. Ventricular pacing is performed from multiple epicardial locations across both the left
and right ventricles, producing a diverse set of paced beats.

The complete dataset comprises recordings from seven different hearts, totaling 380 paced beats.
After removing beats affected by artifacts, 326 beats are retained for analysis. Data from three hearts (309 beats) are used for training and
validation, while data from the remaining four hearts (17 beats with unique pacing locations)
are reserved for testing to assess generalization across subjects. In this dataset, body surface potential measurements are not directly recorded but are simulated from the
experimental epicardial potentials by solving the ECG forward problem using a boundary element
method. The simulated torso potentials are sampled at 192 torso locations and corrupted with
additive Gaussian noise at a signal-to-noise ratio of 20~dB.

In this work, the epicardial potentials serve as the reconstruction target, while the simulated
torso potentials are used as conditioning observations. Although
the paper of~\cite{Ugurlu_pulse} incorporates an explicit forward operator during reconstruction,
we adopt the dataset as paired measurements, consistent with 
geometry-free inverse modeling.

\subsection{Results}
To ensure a fair comparison, we restrict our evaluation to data-driven, geometry-free approaches. Specifically, we compare against neural network models that directly learn the inverse mapping  without incorporating physics-based information or patient-specific geometry. The evaluated baselines include a one-dimensional convolutional neural network (1D-CNN), a long short-term memory (LSTM) network, and a transformer model. The proposed diffusion model and all baselines were trained for 100 epochs using the Adam optimizer with a learning rate of $3\times10^{-4}$ and a batch size of 32. Model capacity was matched by selecting comparable hidden dimensions across architectures. The diffusion model used 100 diffusion timesteps with a quadractic linear noise schedule. All models are trained and tested on the same data splits to ensure a fair comparison.

We report reconstruction performance using a combination of pointwise error and correlation metrics. Specifically, we compute the mean absolute error (MAE) and mean squared error (MSE) between the predicted and ground truth epicardial potentials, which quantify absolute deviations in signal amplitude and penalize large reconstruction errors. We additionally compute the Pearson correlation coefficient (CC) along the temporal dimension. For each heart surface electrode and each beat, the correlation is computed across time, and the resulting coefficients are then averaged across all electrodes and beats to obtain a single summary CC metric. This metric is particularly informative in ECGI, where preserving waveform morphology is clinically important as is exact amplitude matching.

\begin{table}[h]
\centering
\caption{Comparison of reconstruction performance across models.}
\label{tab:model_comparison}
\begin{tabular}{l c c c}
\toprule
Model & Temporal CC ($\uparrow$) & MSE ($\downarrow$) & MAE ($\downarrow$) \\
\midrule
1D-CNN        & 0.77   & 34.89     & 3.79    \\
LSTM          & 0.70   & 47.10  & 4.29    \\
Transformer   & 0.76   & 36.91     & 3.74    \\
\textbf{Diffusion (proposed)} & \textbf{0.78} & \textbf{32.83} & \textbf{3.42}  \\
\bottomrule
\end{tabular}
\end{table}

Table~\ref{tab:model_comparison} compares the reconstruction performance of the proposed diffusion model with 1D-CNN, LSTM, and transformer baselines. The diffusion model achieves the highest temporal Pearson correlation coefficient (0.78) while also obtaining the lowest MSE (32.83) and MAE (3.42) among all evaluated methods. This indicates that the proposed approach more accurately recovers both the temporal morphology and the amplitude of the epicardial signals.

Among the deterministic baselines, the 1D-CNN and Transformer models achieve comparable performance, while the LSTM exhibits lower temporal correlation and higher reconstruction error, suggesting that purely recurrent modeling is less effective in capturing the complex spatiotemporal structure of cardiac electrical activity. The consistent improvement of the diffusion model across all metrics highlights the benefits of the diffusion-based iterative denoising process, which allows progressive refinement of reconstructions under strong conditioning on the observed BSP signals.

\begin{figure}[h]
    \centering
    \includegraphics[width=0.9\linewidth]{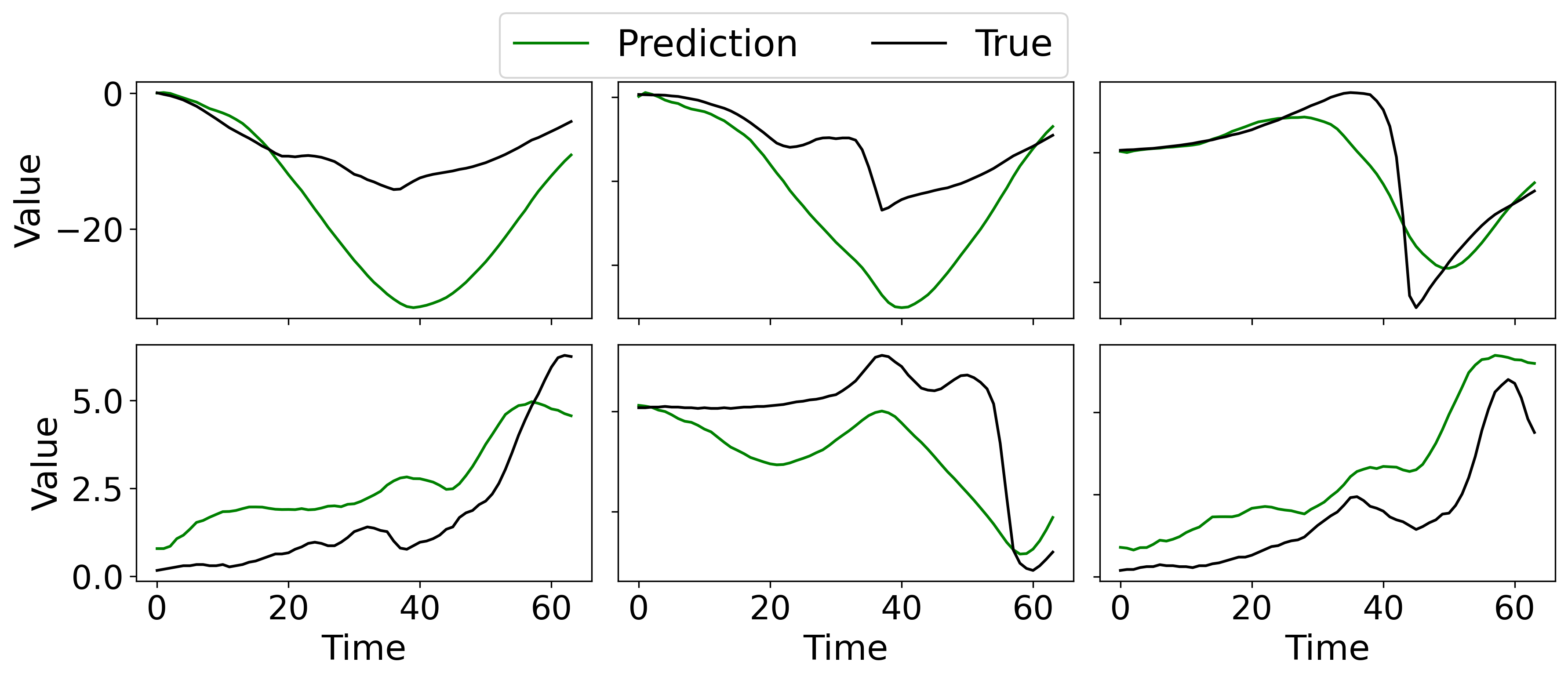}
    \caption{Reconstruction of 6 epicardial point signals.}
    \label{fig:epi_reconst}
\end{figure}

Figure~\ref{fig:epi_reconst} presents representative epicardial reconstructions at six electrode locations selected based on different quantitative criteria. The top row shows electrodes with high temporal correlation to the ground truth, highlighting the model’s ability to recover waveform morphology, while the bottom row shows electrodes with low mean squared error, emphasizing accurate amplitude reconstruction. Across all electrodes, the diffusion model’s predicted mean closely follows the true epicardial signals and captures the dominant temporal patterns.
These qualitative examples complement the quantitative evaluation by illustrating that the proposed method achieves consistent and accurate reconstructions, supporting its effectiveness for the ECGI inverse problem.

Although our diffusion model enables sampling from a learned conditional distribution, the empirical variability observed across samples in this experiment was relatively narrow. This behavior is likely driven by the data generation process, since all body surface measurements were synthesized using a fixed noise level of approximately 20 dB, together with a single forward modeling configuration, which constrains the range of plausible epicardial reconstructions. As a result, the conditional distribution learned by the model is concentrated. We hypothesize that broader uncertainty estimates would emerge if the model were trained under more diverse noise realizations and modeling error patterns, which will be explored in the future.

\section{Conclusion}

This work presents a data-driven conditional diffusion framework for solving the inverse electrocardiography problem in a geometry-free setting. By learning the conditional distribution of heart surface electric potentials given body surface measurements, the proposed approach accounts for both the ill-posed nature of ECGI and the inherent uncertainty of the inverse mapping, while avoiding the need for patient-specific anatomical modeling.

Experimental results demonstrate that the diffusion-based model consistently outperforms deterministic baselines, including 1D-CNN, LSTM, and transformer architectures, across all  metrics. In particular, the proposed method achieves higher temporal correlation with ground truth signals while also reducing pointwise reconstruction errors, indicating improved recovery of both waveform morphology and signal amplitude.

This study emphasizes diffusion models as a suitable probabilistic modeling tool for ECGI, offering a powerful alternative to traditional deterministic approaches, especially in settings where anatomical information is missing. While the empirical uncertainty observed in the present experiments is relatively concentrated due to the controlled data generation process, the proposed framework provides a natural foundation for uncertainty-aware ECGI reconstruction. Future work will focus on training under diverse noise levels and more heterogeneous forward-model perturbations to better capture the full distribution of physiologically plausible epicardial reconstructions and improve uncertainty calibration under realistic clinical conditions.

\bibliographystyle{ieeetr}
\bibliography{references}

\end{document}